\pdfoutput=1

\documentclass[11pt]{article}

\usepackage{authblk}
\usepackage{acl}

\usepackage{times}
\usepackage{latexsym}

\usepackage{amsmath}
\usepackage{booktabs}
\usepackage[hang,flushmargin]{footmisc} 

\usepackage[T1]{fontenc}

\usepackage[utf8]{inputenc}

\usepackage{microtype}

\usepackage[disable]{todonotes}
\presetkeys{todonotes}{inline}{}
\newcommand{\respace}{\vspace*{-.1cm}}

\author[*123]{\textbf{Rajiv Movva}}
\author[*2]{\textbf{Jinhao Lei}}
\author[23]{\textbf{Shayne Longpre}}
\author[2]{\textbf{Ajay Gupta}}
\author[2]{\textbf{Chris DuBois}}
\affil[*]{Equal Contribution}
\affil[1]{Cornell Tech, rm868@cornell.edu}
\affil[2]{Apple, \{jlei2, ajay\_gupta2, cdubois\}@apple.com}
\affil[3]{Massachusetts Institute of Technology, slongpre@mit.edu}

\title{Combining Compressions for Multiplicative Size Scaling on Natural Language Tasks}

\begin{document}
\maketitle

\begin{abstract}
\respace
Quantization, knowledge distillation, and magnitude pruning are among the most popular methods for neural network compression in NLP.
Independently, these methods reduce model size and can accelerate inference, but their relative benefit and combinatorial interactions have not been rigorously studied.
For each of the eight possible subsets of these techniques, we compare accuracy vs.~model size tradeoffs across six BERT architecture sizes and eight GLUE tasks.
We find that quantization and distillation consistently provide greater benefit than pruning. 
Surprisingly, except for the pair of pruning and quantization, using multiple methods together rarely yields diminishing returns. 
Instead, we observe complementary and super-multiplicative reductions to model size.
Our work quantitatively demonstrates that combining compression methods can synergistically reduce model size, and that practitioners should prioritize (1) quantization, (2) knowledge distillation, and (3) pruning to maximize accuracy vs. model size tradeoffs.


\end{abstract}

\respace
\section{Introduction}
\respace

As increasingly large models dominate Natural Language Processing (NLP) benchmarks, model compression techniques have grown in popularity  \citep{gupta2020compression, rogers2020primer, ganesh2021compressing}.
For example, quantization \citep{shen2020q, zafrir2019q8bert, jacob2018quantization} lowers bit precision of network weights to reduce memory usage and accelerate inference \citep{krashinsky_nvidia_2020}.
Knowledge distillation (\textsc{KD}; \citet{hinton2015distilling}), which trains a student neural network using the logits (or representations) of a teacher network, is used widely to transfer knowledge to smaller models \citep{sanh2019distilbert, jiao2020tinybert, sun2019patient, sun-etal-2020-mobilebert}.
Pruning identifies weights which can be omitted at test time without significantly degrading performance.
Some pruning methods remove individual weights according to magnitudes or other heuristics \citep{gordon_compressing_2020, NEURIPS2020_b6af2c97, sanh_movement_2020}, while others remove structured blocks of weights or entire attention heads \citep{wang_structured_2020, hou_dynabert_2020, voita_analyzing_2019, michel_are_2019}.

Recent work has begun combining these compression methods for improved results.
\citet{sanh_movement_2020}, \citet{zhang2020ternarybert}, and \citet{bai-etal-2021-binarybert} have used knowledge distillation with pruning or low-bit quantization to fine-tune BERT.
As practitioners look to combine methods more generally, new research is needed to compare their empirical value and study interactions.
This work addresses the questions: (1) Which popular compression methods or combinations of methods are usually most effective? (2) When combining methods, are their benefits complementary or diminishing?

We address these questions by computing accuracy vs. model size tradeoff curves for six pre-trained BERT sizes fine-tuned on eight GLUE tasks \citep{wang2019glue}, applying each of eight possible subsets of quantization-aware-training (QAT), knowledge distillation (KD), and magnitude pruning (MP). 
Our main findings are as follows:
\begin{enumerate}\itemsep0em
  \item When methods are applied independently, \textsc{QAT} yields best accuracy-compression tradeoffs, followed by \textsc{KD} and then \textsc{MP}. 
  
  \item Strikingly, we observe no diminishing returns when combining \textsc{KD} with \textsc{QAT} or \textsc{MP}.
  Instead, \textsc{KD} mitigates the loss in accuracy caused by either method, thereby super-multiplicatively reducing model size.
  
  \item When used together, \textsc{QAT} and \textsc{MP} amplify each other's individual accuracy losses. However, combining all three methods (\textit{i.e.}, also using KD) preserves accuracy, allowing 18x and 11x compression for BERT-\textsc{Large} and \textsc{Base} respectively. 
\end{enumerate}

\respace
\section{Methods}
\respace

In our work, we study three common model compressions: quantization-aware-training, knowledge distillation, and magnitude pruning. 
We prioritize performant, broadly applicable approaches with accessible implementations, so our findings are most useful to practictioners.
Hyperparameters and additional method details are in Appendices A \& C.

\paragraph{BERT Architecture Sizes.} We test each compression combination across six different BERT architecture sizes, seeing as they may have different compressibilities.
These pretrained models are taken from \citet{turc_well-read_2019}: \textsc{Large} (367 million params), \textsc{Base} (134M), \textsc{Medium} (57M), \textsc{Small} (45M), \textsc{Mini} (19M), \textsc{Tiny} (8M). 
Including a range of sizes makes our findings relevant to practitioners or deployment settings without resources for architectures like \textsc{Large} or \textsc{Base}.
Additionally, using smaller model sizes is a practical baseline to compare our compression methods against.

\paragraph{Quantization-Aware-Training (\textsc{QAT}).}
While most neural networks use 32-bit floats for weights and activations, recent work has shown promise for lower precisions.
16-bit floats cause no accuracy loss for most architectures \citep{das_mixed_2018}, and \citet{zafrir2019q8bert} show that, with quantization-aware-training (\textsc{QAT}), 8-bit integer (INT8) BERT mostly preserves GLUE accuracy. 
The INT8 model is nearly 4x smaller, and can achieve 2.4-4.0x inference acceleration with appropriate hardware \cite{kim_i-bert_2021}. 
As lower precisions harm accuracy significantly \citep{shen2020q}, we use an 8-bit BERT with the \textsc{QAT} scheme described by \citet{zafrir2019q8bert}, recapped in the Appendix. 


\paragraph{Knowledge Distillation (\textsc{KD}).} 
In \textsc{KD}, we fine-tune a small student model by optimizing its weights to mimic the outputs of a teacher model. 
We use a common, simple variant of \textsc{KD}, emulating \citet{turc_well-read_2019}: we use a BERT-\textsc{Large} fine-tuned for three epochs on the GLUE task as the teacher, and the student is trained to minimize KL-divergence between its predicted probabilities and the teacher's. 

To further improve the utility of KD, we adopt \citet{jiao2020tinybert}'s approach of data augmentation (DA) for GLUE training datasets. 
This technique helps for all tasks, especially the smaller ones (\textit{e.g.} \textsc{MRPC}, \textsc{RTE}). 
Each example is copied 10, 20, or 30 times (more copies for smaller tasks), and each copy has some of its words replaced with synonyms (\textit{i.e.} words with closest GLoVe embeddings). 
Many of the copies have altered meanings, but the teacher is able to adapt by making different predictions.
Before running all of our experiments, we ran a few trials (on \textsc{MRPC} and \textsc{QNLI}) to confirm that DA helped with distillation but not without, and so we only used DA for the KD experiments.


\paragraph{Magnitude Pruning (\textsc{MP}).}

Several pruning methods have been used in NLP \citep{hoefler_sparsity_2021}. 
We use unstructured weight pruning, which can achieve higher sparsities than structured pruning \citep{renda_comparing_2020}, and has comparatively standard implementations.

Magnitude pruning masks the weights with lowest magnitudes to achieve a target sparsity. 
As in \citet{sanh_movement_2020}, we iteratively prune weights at a linear schedule during training after some warmup steps. 
\citet{sanh_movement_2020} also propose \textit{movement pruning}, in which weights are pruned according to their gradients during fine-tuning. 
We found that movement pruning performs worse than magnitude at moderate sparsities (40-60\%), when accuracy is retained (corroborated by \citet{sanh_movement_2020}). 
As we target accuracy-preserving pruning, we use magnitude pruning for the experiments in this work. 
We prune either 40\% or 60\% of encoder weights only, as pruning embedding weights significantly damages accuracy \citep{yu_playing_2020}.
\respace
\section{Results}
\respace

\begin{figure*}[!ht]
    \respace
    \respace
    \centering
    \includegraphics[width=\textwidth]{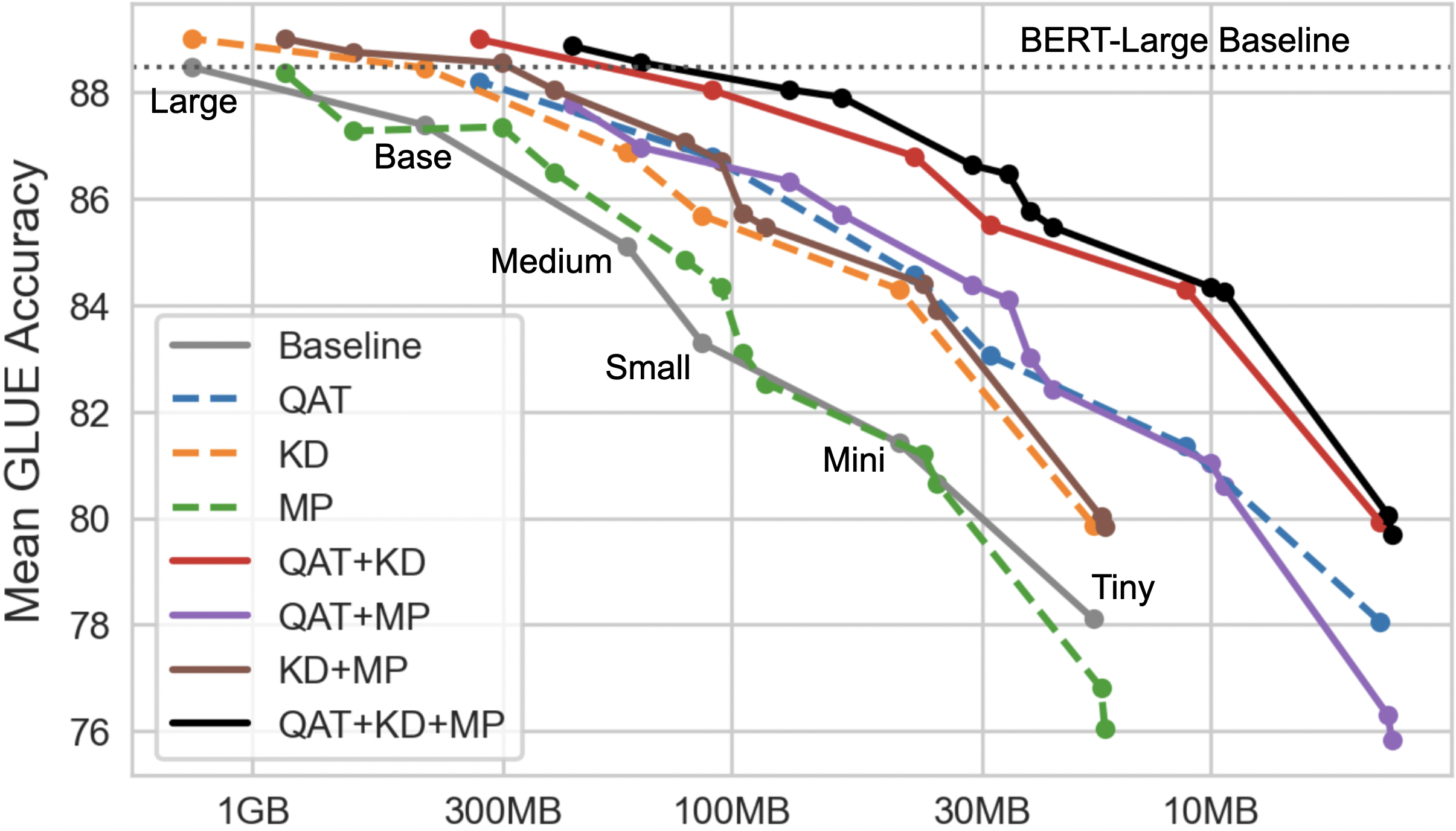}
    \caption{Mean GLUE accuracy vs. \textbf{decreasing} model size, with curves plotted for each compression combination. The different points for each curve represent the different BERT architecture sizes, from \textsc{Large} down to \textsc{Tiny}.}
    \label{fig:meanglue}
    \respace
    \respace
\end{figure*}

\paragraph{Experiments} For six BERT architecture sizes and eight GLUE tasks\footnote{\,We excluded CoLA and WNLI to reduce experimental burden and due to issues with WNLI \citep{wang-etal-2019-tell}.}, we tested every possible subset of compression methods: no compression (Baseline), \textsc{QAT}, \textsc{KD}, \textsc{MP}, \textsc{QAT+KD}, \textsc{QAT+MP}, \textsc{KD+MP}, and \textsc{QAT+KD+MP}.
For each of $576$ experiment settings, we log the max GLUE development set accuracy across twelve hyperparameter configurations and five repetitions of each configuration.
In Figure \ref{fig:meanglue}, we plot mean GLUE accuracy across all eight tasks on the $y$-axis against decreasing model size on the $x$-axis, for each compression combination. 
The curves without pruning include six points, one for each architecture size from \textsc{Large} to \textsc{Tiny}. 
With pruning, there are twice as many points, as each architecture is pruned to either $40\%$ or $60\%$ encoder sparsity. 
Results split by task are available in Appendix Figure \ref{fig:allglue}.

\paragraph{Individually, \textsc{QAT} and \textsc{KD} are most effective.} 
For all architectures, \textsc{QAT} (blue) reduces model size by 4$\times$ while minimally reducing accuracy: the largest drop is $-0.6\%$ for BERT-\textsc{Base} (supporting \citet{zafrir2019q8bert}), while other architecture sizes are nearly unaffected. 
\textsc{KD} (orange) does not reduce a given architecture's size, but instead yields a consistent boost to accuracy, especially for smaller architectures.
This upward shift on the accuracy-model size curve means that larger models can be downsized more effectively: \textit{e.g.}, \textsc{Large}'s baseline accuracy is matched by the \textsc{KD} version of \textsc{Base}, and \textsc{KD} \textsc{Small} outperforms baseline \textsc{Medium}. 
Compared to \textsc{QAT} or \textsc{KD}, \textsc{MP} (green) is only modestly helpful. Typically, $40\%$ of encoder weights can be removed without much impact, but pruning $60\%$ (\textit{i.e.}, the second point in each set of two) degrades accuracy.
Also, removing $40\%$ of encoder weights corresponds to a $<40\%$ model size reduction because we do not prune embedding weights.
Therefore, while \textsc{MP} can be helpful for \textsc{Large} and \textsc{Base}, it cannot significantly compress small architectures, which have a higher percentage of their weights in embeddings.

\paragraph{Used together, \textsc{KD} mitigates accuracy losses from both \textsc{QAT} and \textsc{MP}.}
Moving to combinations of compression methods, we find that the most successful pair combines \textsc{QAT} and \textsc{KD} (red), which yields \textsc{QAT}'s 4$\times$ memory reduction while retaining the improved accuracy over baseline from \textsc{KD}.
Meanwhile, \textsc{QAT}+\textsc{MP} (purple) does poorly: though \textsc{Large} and \textsc{Base} can prune $40\%$ of weights while retaining accuracy, when they are pruned \textit{and} quantized, they have lower accuracy than when they are only quantized. 
This result suggests that pruning specifically damages accuracy with quantization: practitioners should expect additive (or worse) accuracy degradation when combining \textsc{QAT} and \textsc{MP}.
On the other hand, with \textsc{KD}+\textsc{MP} (brown), $40\%$ weights can be pruned while retaining the accuracy boost from \textsc{KD}, for all architectures. 
Thus, \textsc{KD} \textit{mitigates} accuracy losses from both \textsc{MP} and \textsc{QAT}.
This result still holds when we combine all three methods (black).
With \textsc{QAT}+\textsc{KD}+\textsc{MP}, $40\%$ of encoder weights for \textsc{Large} and $60\%$ for \textsc{Base} (and smaller) can be removed while matching the accuracy of \textsc{QAT}+\textsc{KD}.
In our experiments, \textsc{KD} completely mitigates the compounding losses from \textsc{QAT}+\textsc{MP} and even \textit{improves} accuracy.
\textsc{KD} enables deeper compression when practitioners combine methods.

\begin{table*}[!h]
\respace
\respace
\centering
    \begin{tabular}{lcccccccc}
    \toprule
    {} & \textbf{Size} & \textbf{\textcolor{blue}{\textsc{QAT}}} &   \textbf{\textcolor{orange}{\textsc{KD}}} &  \textbf{\textcolor{ForestGreen}{\textsc{MP}}} &  \textbf{\textcolor{blue}{\textsc{QAT}}+\textcolor{orange}{\textsc{KD}}} &  \textbf{\textcolor{blue}{\textsc{QAT}}+\textcolor{ForestGreen}{\textsc{MP}}} &  \textbf{\textcolor{orange}{\textsc{KD}}+\textcolor{ForestGreen}{\textsc{MP}}} &  \textbf{\textcolor{blue}{\textsc{QAT}}+\textcolor{orange}{\textsc{KD}}+\textcolor{ForestGreen}{\textsc{MP}}} \\ 
    \midrule 
    \textsc{Large}  &  1341 &  $3.6\times$ &  $3.4\times$ &     $1.7\times$ &     $12.6\times$ &          $3.9\times$ &        $5.8\times$ &            $18.1\times$ \\
    \textsc{Base}   & 438 &  $2.6\times$ &  $1.8\times$ &     $1.7\times$ &      $5.9\times$ &          $3.1\times$ &        $2.8\times$ &            $11.1\times$ \\
    \textsc{Medium} & 166 &  $2.9\times$ &  $2.0\times$ &     $1.2\times$ &     $8.0\times$ &          $4.0\times$ &        $3.5\times$ &            $14.1\times$ \\
    \textsc{Small}  & 115 &  $3.6\times$ &  $2.4\times$ &     $1.2\times$ &      $9.4\times$ &          $4.7\times$ &        $3.6\times$ &            $14.1\times$ \\
    \textsc{Mini}   & 45 &   $3.7\times$ &  $1.2\times$ &     $1.1\times$ &      $4.7\times$ &          $3.4\times$ &        $1.8\times$ &             $7.0\times$ \\ 
    \textsc{Tiny}   & 18 &  $3.4\times$ &  $1.0\times$ &     $0.7\times$ &      $4.0\times$ &          $2.2\times$ &        $1.0\times$ &             $3.9\times$ \\ \bottomrule
    \end{tabular}
    \caption{Ratios, averaged across all GLUE tasks, measuring the maximum possible size reduction factor of a certain architecture while within $0.5\%$ of baseline accuracy. Uncompressed sizes are listed in \textbf{megabytes (MB)}.}
    \label{tbl:ratios}
    \respace
    \respace
\end{table*}

\paragraph{Combining methods yields \textit{super-multiplicative} compression ratios.} 
Building on our qualitative findings, we were interested in quantitative estimates for how much each method allows us to compress each architecture size.
So, we compute \textit{compression ratios}, \textit{i.e.}, the maximal size reduction factor possible while preserving accuracy to within $0.5\%$ of baseline.
For example, baseline \textsc{Large} (1341 MB) yields $92.8\%$ accuracy on QNLI, while \textsc{Base} with \textsc{QAT}, \textsc{KD}, and $40\%$ \textsc{MP} (76 MB) is the smallest model within $0.5\%$ of that, at $92.6\%$ accuracy. 
Therefore, BERT-\textsc{Large}'s compression ratio on QNLI is $\frac{1341}{76} = 17.6\times$, and averaging this value across tasks yields a net ratio of $18.1\times$ (top-right, Table \ref{tbl:ratios}). We similarly compute ratios for all architectures and compression combinations.



As before, QAT usually retains accuracy and yields a $4\times$ size reduction. 
However, because there are a few tasks for which QAT causes a $>0.5\%$ accuracy drop\footnote{\,Those tasks would be considered to have $1\times$ compression.}, the task-averaged compression ratios for \textsc{Large} and \textsc{Base} of $3.6\times$ and $2.6\times$.
\textsc{KD} also has high mean compression ratios, because it often boosts small architectures' accuracies to match larger baseline architectures. 
MP yields $1.7\times$ compression (from $40\%$ pruning) for \textsc{Large} and \textsc{Base}, and even less for the smaller architectures.


When combined, we observe synergistic compression between \textsc{QAT} and \textsc{KD}. 
We might expect strong diminishing returns from combining methods, but even in their absence, we would expect independent compression ratios to multiply.
Strikingly, though, we often see \textit{super-multiplicative} model size reductions with \textsc{QAT}+\textsc{KD}: \textit{e.g.} \textsc{Base}, $5.9 > 2.6 \cdot 1.8 = 4.7$; \textsc{Medium}: $8.0 > 2.9 \cdot 2.0 = 5.8$. 
For \textsc{KD}+\textsc{MP}, the ratios multiply for \textsc{Large} and \textsc{Base}. 
Also, while pruning was originally ineffective for \textsc{Medium} and smaller, \textsc{KD}+\textsc{MP} appears to make pruning more effective, again with super-multiplicative compression (e.g. for Medium: $3.5 > 2.0 \cdot 1.2 = 2.4$).
These findings show that \textsc{KD} mitigates the drop in accuracy from quantization \citep{zhang2020ternarybert} and pruning \citep{sanh_movement_2020}, supporting qualitative findings from prior literature. 
This mitigation effect, combined with the accuracy boost that \textsc{KD} provides, yields joint compressions that are more effective than the sum of their parts.
Remarkably, with all three compressions, we still see super-multiplicative scaling for \textsc{Base} and smaller (\textit{e.g.}, \textsc{Base}: $11.1 > 2.6 \cdot 1.8 \cdot 1.7 = 8.0$; \textsc{Medium}: $14.1 > 2.9\cdot2.0\cdot1.2 = 7.0$). 
At $18.1\times$, \textsc{Large} is most compressible, but actually has a sub-multiplicative ratio, perhaps because the individual compressions already work very well.

\respace
\section{Discussion and Future Work}
\respace

We examine the relative benefits and interactions of three widely-used compression methods in NLP: knowledge distillation, INT8 quantization-aware-training, and magnitude pruning. 
Though multiple techniques are increasingly used simultaneously, little work has studied the empirical interactions between them.

Across six architecture sizes and eight GLUE tasks, we find that INT8 quantization is most beneficial, and combining quantization with KD mitigates any of its occasional accuracy drops. 
Quantization and pruning yield diminishing returns, with the two methods exacerbating accuracy losses when used together.
However, when all three methods are combined, KD restores accuracy above baseline, yielding $18\times$ and $11\times$ net compression for BERT-\textsc{Large} and \textsc{Base}. 
We quantitatively confirm that KD's model size improvements are complementary (often super-multiplicative) with quantization and pruning. 
Given these benefits, using larger, more recent GLUE winners as KD teachers may yield further gains (\textit{e.g.} ERNIE, \citet{sun_ernie_2021}). 
We also hope that our observed benefits inspire authors of new compression techniques to evaluate complementarity with existing methods.


In future work, we hope to measure accuracy vs. inference time tradeoffs, and compare these results to our findings on model size. 
Though there is work on accelerated inference with INT8 quantization \citep{kim2021bert} or pruning \citep{nvidia2021}, it is not clear how these speedups stack. 
Profiling compression combinations on specialized hardware would be an informative avenue to explore next.




\section*{Broader Impacts}

As NLP models dramatically scale in size, they rely increasingly on specialized hardware (\textit{e.g.} GPUs, TPUs) to train and deploy them. 
The manufacturing and energy consumption involved in the usage of such devices imposes a significant carbon footprint \citep{strubell_energy_2019, gupta_chasing_2021}. 
Model compression is part of the broader movement towards "Green AI", in which researchers develop more energy-efficient models with similar task accuracies in order to reduce usage of compute-hungry hardware \citep{schwartz_green_2020}.

By careful empirical study of how to optimally combine compression methods, we believe our work takes further steps towards Green AI. 
In particular, rather than proposing a single architecture that achieves a specific tradeoff, we equip practitioners with a set of principles to apply depending on their needs, hopefully increasing uptake of at least some subset of compression methods. 
Though the present study itself required results from thousands of training runs, there is significant potential for this work to pay off if models in widely-used deployment settings (speech-to-text, search, \textit{etc.}) draw from our findings.

Though model compression methods can dramatically mitigate deep learning's carbon footprint, they may also create opportunities for harm \citep{suresh_framework_2021}.
Specifically, recent work shows that pruning damages accuracy for minority classes in the training dataset \citep{hooker_characterising_2020}, and that pruning may change model behavior even when accuracy is preserved \citep{movva_dissecting_2020}. 
Such predictive disparities can lead to algorithmic harms: \textit{e.g.,} \textit{representational} harms for language models, or \textit{allocational} harms for certain downstream task predictors \cite{blodgett_language_2020}. 
More work is needed to systematically characterize the relationship between compression and algorithmic harms.



\bibliography{anthology,custom}

\begin{thebibliography}{42}
\expandafter\ifx\csname natexlab\endcsname\relax\def\natexlab#1{#1}\fi

\bibitem[{Bai et~al.(2021)Bai, Zhang, Hou, Shang, Jin, Jiang, Liu, Lyu, and
  King}]{bai-etal-2021-binarybert}
Haoli Bai, Wei Zhang, Lu~Hou, Lifeng Shang, Jin Jin, Xin Jiang, Qun Liu,
  Michael Lyu, and Irwin King. 2021.
\newblock \href {https://doi.org/10.18653/v1/2021.acl-long.334}
  {{B}inary{BERT}: Pushing the limit of {BERT} quantization}.
\newblock In \emph{Proceedings of the 59th Annual Meeting of the Association
  for Computational Linguistics and the 11th International Joint Conference on
  Natural Language Processing (Volume 1: Long Papers)}, pages 4334--4348,
  Online. Association for Computational Linguistics.

\bibitem[{Blodgett et~al.(2020)Blodgett, Barocas, Daumé~III, and
  Wallach}]{blodgett_language_2020}
Su~Lin Blodgett, Solon Barocas, Hal Daumé~III, and Hanna Wallach. 2020.
\newblock \href {http://arxiv.org/abs/2005.14050} {Language ({Technology}) is
  {Power}: {A} {Critical} {Survey} of "{Bias}" in {NLP}}.
\newblock \emph{arXiv:2005.14050 [cs]}.
\newblock ArXiv: 2005.14050.

\bibitem[{Chen et~al.(2020)Chen, Frankle, Chang, Liu, Zhang, Wang, and
  Carbin}]{NEURIPS2020_b6af2c97}
Tianlong Chen, Jonathan Frankle, Shiyu Chang, Sijia Liu, Yang Zhang, Zhangyang
  Wang, and Michael Carbin. 2020.
\newblock \href
  {https://proceedings.neurips.cc/paper/2020/file/b6af2c9703f203a2794be03d443af2e3-Paper.pdf}
  {The lottery ticket hypothesis for pre-trained bert networks}.
\newblock In \emph{Advances in Neural Information Processing Systems},
  volume~33, pages 15834--15846. Curran Associates, Inc.

\bibitem[{Das et~al.(2018)Das, Mellempudi, Mudigere, Kalamkar, Avancha,
  Banerjee, Sridharan, Vaidyanathan, Kaul, Georganas, Heinecke, Dubey, Corbal,
  Shustrov, Dubtsov, Fomenko, and Pirogov}]{das_mixed_2018}
Dipankar Das, Naveen Mellempudi, Dheevatsa Mudigere, Dhiraj~D. Kalamkar,
  Sasikanth Avancha, Kunal Banerjee, Srinivas Sridharan, Karthik Vaidyanathan,
  Bharat Kaul, Evangelos Georganas, Alexander Heinecke, Pradeep Dubey,
  Jes{\'{u}}s Corbal, Nikita Shustrov, Roman Dubtsov, Evarist Fomenko, and
  Vadim~O. Pirogov. 2018.
\newblock \href {http://arxiv.org/abs/1802.00930} {Mixed precision training of
  convolutional neural networks using integer operations}.
\newblock \emph{CoRR}, abs/1802.00930.

\bibitem[{Devlin et~al.(2018)Devlin, Chang, Lee, and
  Toutanova}]{devlin_bert_2018}
Jacob Devlin, Ming{-}Wei Chang, Kenton Lee, and Kristina Toutanova. 2018.
\newblock \href {http://arxiv.org/abs/1810.04805} {{BERT:} pre-training of deep
  bidirectional transformers for language understanding}.
\newblock \emph{CoRR}, abs/1810.04805.

\bibitem[{Ganesh et~al.(2021)Ganesh, Chen, Lou, Khan, Yang, Sajjad, Nakov,
  Chen, and Winslett}]{ganesh2021compressing}
Prakhar Ganesh, Yao Chen, Xin Lou, Mohammad~Ali Khan, Yin Yang, Hassan Sajjad,
  Preslav Nakov, Deming Chen, and Marianne Winslett. 2021.
\newblock Compressing large-scale transformer-based models: A case study on
  bert.
\newblock \emph{Transactions of the Association for Computational Linguistics},
  9:1061--1080.

\bibitem[{Gordon et~al.(2020)Gordon, Duh, and
  Andrews}]{gordon_compressing_2020}
Mitchell~A. Gordon, Kevin Duh, and Nicholas Andrews. 2020.
\newblock \href {http://arxiv.org/abs/2002.08307} {Compressing {BERT}:
  {Studying} the {Effects} of {Weight} {Pruning} on {Transfer} {Learning}}.
\newblock \emph{arXiv:2002.08307 [cs]}.
\newblock ArXiv: 2002.08307.

\bibitem[{Gupta and Agrawal(2020)}]{gupta2020compression}
Manish Gupta and Puneet Agrawal. 2020.
\newblock Compression of deep learning models for text: A survey.
\newblock \emph{arXiv preprint arXiv:2008.05221}.

\bibitem[{Gupta et~al.(2021)Gupta, Kim, Lee, Tse, Lee, Wei, Brooks, and
  Wu}]{gupta_chasing_2021}
Udit Gupta, Young~Geun Kim, Sylvia Lee, Jordan Tse, Hsien-Hsin~S. Lee, Gu-Yeon
  Wei, David Brooks, and Carole-Jean Wu. 2021.
\newblock \href {https://doi.org/10.1109/HPCA51647.2021.00076} {Chasing
  {Carbon}: {The} {Elusive} {Environmental} {Footprint} of {Computing}}.
\newblock In \emph{2021 {IEEE} {International} {Symposium} on
  {High}-{Performance} {Computer} {Architecture} ({HPCA})}, pages 854--867.
\newblock ISSN: 2378-203X.

\bibitem[{Hinton et~al.(2015)Hinton, Vinyals, and Dean}]{hinton2015distilling}
Geoffrey Hinton, Oriol Vinyals, and Jeff Dean. 2015.
\newblock Distilling the knowledge in a neural network.
\newblock \emph{arXiv preprint arXiv:1503.02531}.

\bibitem[{Hoefler et~al.(2021)Hoefler, Alistarh, Ben-Nun, Dryden, and
  Peste}]{hoefler_sparsity_2021}
Torsten Hoefler, Dan Alistarh, Tal Ben-Nun, Nikoli Dryden, and Alexandra Peste.
  2021.
\newblock \href {http://arxiv.org/abs/2102.00554} {Sparsity in {Deep}
  {Learning}: {Pruning} and growth for efficient inference and training in
  neural networks}.
\newblock \emph{arXiv:2102.00554 [cs]}.
\newblock ArXiv: 2102.00554.

\bibitem[{Hooker et~al.(2020)Hooker, Moorosi, Clark, Bengio, and
  Denton}]{hooker_characterising_2020}
Sara Hooker, Nyalleng Moorosi, Gregory Clark, Samy Bengio, and Emily Denton.
  2020.
\newblock \href {http://arxiv.org/abs/2010.03058} {Characterising {Bias} in
  {Compressed} {Models}}.
\newblock \emph{arXiv:2010.03058 [cs]}.
\newblock ArXiv: 2010.03058.

\bibitem[{Hou et~al.(2020)Hou, Huang, Shang, Jiang, Chen, and
  Liu}]{hou_dynabert_2020}
Lu~Hou, Zhiqi Huang, Lifeng Shang, Xin Jiang, Xiao Chen, and Qun Liu. 2020.
\newblock \href
  {https://proceedings.neurips.cc/paper/2020/file/6f5216f8d89b086c18298e043bfe48ed-Paper.pdf}
  {Dynabert: Dynamic bert with adaptive width and depth}.
\newblock In \emph{Advances in Neural Information Processing Systems},
  volume~33, pages 9782--9793. Curran Associates, Inc.

\bibitem[{Jacob et~al.(2017)Jacob, Kligys, Chen, Zhu, Tang, Howard, Adam, and
  Kalenichenko}]{jacob_quantization_2017}
Benoit Jacob, Skirmantas Kligys, Bo~Chen, Menglong Zhu, Matthew Tang, Andrew
  Howard, Hartwig Adam, and Dmitry Kalenichenko. 2017.
\newblock \href {http://arxiv.org/abs/1712.05877} {Quantization and {Training}
  of {Neural} {Networks} for {Efficient} {Integer}-{Arithmetic}-{Only}
  {Inference}}.
\newblock \emph{arXiv:1712.05877 [cs, stat]}.
\newblock ArXiv: 1712.05877.

\bibitem[{Jacob et~al.(2018)Jacob, Kligys, Chen, Zhu, Tang, Howard, Adam, and
  Kalenichenko}]{jacob2018quantization}
Benoit Jacob, Skirmantas Kligys, Bo~Chen, Menglong Zhu, Matthew Tang, Andrew
  Howard, Hartwig Adam, and Dmitry Kalenichenko. 2018.
\newblock Quantization and training of neural networks for efficient
  integer-arithmetic-only inference.
\newblock In \emph{2018 IEEE/CVF Conference on Computer Vision and Pattern
  Recognition (CVPR)}, pages 2704--2713. IEEE Computer Society.

\bibitem[{Jiao et~al.(2020)Jiao, Yin, Shang, Jiang, Chen, Li, Wang, and
  Liu}]{jiao2020tinybert}
Xiaoqi Jiao, Yichun Yin, Lifeng Shang, Xin Jiang, Xiao Chen, Linlin Li, Fang
  Wang, and Qun Liu. 2020.
\newblock Tinybert: Distilling bert for natural language understanding.
\newblock In \emph{Proceedings of the 2020 Conference on Empirical Methods in
  Natural Language Processing: Findings}, pages 4163--4174.

\bibitem[{Kim et~al.(2021{\natexlab{a}})Kim, Gholami, Yao, Mahoney, and
  Keutzer}]{kim_i-bert_2021}
Sehoon Kim, Amir Gholami, Zhewei Yao, Michael~W. Mahoney, and Kurt Keutzer.
  2021{\natexlab{a}}.
\newblock \href {http://arxiv.org/abs/2101.01321} {I-{BERT}: {Integer}-only
  {BERT} {Quantization}}.
\newblock \emph{arXiv:2101.01321 [cs]}.
\newblock ArXiv: 2101.01321.

\bibitem[{Kim et~al.(2021{\natexlab{b}})Kim, Gholami, Yao, Mahoney, and
  Keutzer}]{kim2021bert}
Sehoon Kim, Amir Gholami, Zhewei Yao, Michael~W Mahoney, and Kurt Keutzer.
  2021{\natexlab{b}}.
\newblock I-bert: Integer-only bert quantization.
\newblock \emph{arXiv preprint arXiv:2101.01321}.

\bibitem[{Krashinsky et~al.(2020)Krashinsky, Giroux, Jones, Stam, and
  Ramaswamy}]{krashinsky_nvidia_2020}
Ronny Krashinsky, Oliver Giroux, Stephen Jones, Nick Stam, and Sridhar
  Ramaswamy. 2020.
\newblock \href
  {https://developer.nvidia.com/blog/nvidia-ampere-architecture-in-depth/}
  {{NVIDIA} {Ampere} {Architecture} {In}-{Depth}}.

\bibitem[{Michel et~al.(2019)Michel, Levy, and Neubig}]{michel_are_2019}
Paul Michel, Omer Levy, and Graham Neubig. 2019.
\newblock \href {http://arxiv.org/abs/1905.10650} {Are {Sixteen} {Heads}
  {Really} {Better} than {One}?}
\newblock \emph{arXiv:1905.10650 [cs]}.
\newblock ArXiv: 1905.10650.

\bibitem[{Movva and Zhao(2020)}]{movva_dissecting_2020}
Rajiv Movva and Jason~Y. Zhao. 2020.
\newblock \href {http://arxiv.org/abs/2009.13270} {Dissecting {Lottery}
  {Ticket} {Transformers}: {Structural} and {Behavioral} {Study} of {Sparse}
  {Neural} {Machine} {Translation}}.
\newblock \emph{arXiv:2009.13270 [cs, stat]}.
\newblock ArXiv: 2009.13270.

\bibitem[{Pennington et~al.(2014)Pennington, Socher, and
  Manning}]{pennington-etal-2014-glove}
Jeffrey Pennington, Richard Socher, and Christopher Manning. 2014.
\newblock \href {https://doi.org/10.3115/v1/D14-1162} {{G}lo{V}e: Global
  vectors for word representation}.
\newblock In \emph{Proceedings of the 2014 Conference on Empirical Methods in
  Natural Language Processing ({EMNLP})}, pages 1532--1543, Doha, Qatar.
  Association for Computational Linguistics.

\bibitem[{Pool et~al.(2021)Pool, Sawarkar, and Rodge}]{nvidia2021}
Jeff Pool, Abhishek Sawarkar, and Jay Rodge. 2021.
\newblock \href
  {https://developer.nvidia.com/blog/accelerating-inference-with-sparsity-using-ampere-and-tensorrt/}
  {Accelerating inference with sparsity using the nvidia ampere architecture
  and nvidia tensorrt}.
\newblock \emph{NVIDIA Developer Blog}.

\bibitem[{Renda et~al.(2020)Renda, Frankle, and Carbin}]{renda_comparing_2020}
Alex Renda, Jonathan Frankle, and Michael Carbin. 2020.
\newblock \href {http://arxiv.org/abs/2003.02389} {Comparing {Rewinding} and
  {Fine}-tuning in {Neural} {Network} {Pruning}}.
\newblock \emph{arXiv:2003.02389 [cs, stat]}.
\newblock ArXiv: 2003.02389.

\bibitem[{Rogers et~al.(2020)Rogers, Kovaleva, and
  Rumshisky}]{rogers2020primer}
Anna Rogers, Olga Kovaleva, and Anna Rumshisky. 2020.
\newblock A primer in bertology: What we know about how bert works.
\newblock \emph{Transactions of the Association for Computational Linguistics},
  8:842--866.

\bibitem[{Sanh et~al.(2019)Sanh, Debut, Chaumond, and
  Wolf}]{sanh2019distilbert}
Victor Sanh, Lysandre Debut, Julien Chaumond, and Thomas Wolf. 2019.
\newblock Distilbert, a distilled version of bert: smaller, faster, cheaper and
  lighter.
\newblock \emph{arXiv preprint arXiv:1910.01108}.

\bibitem[{Sanh et~al.(2020)Sanh, Wolf, and Rush}]{sanh_movement_2020}
Victor Sanh, Thomas Wolf, and Alexander Rush. 2020.
\newblock \href
  {https://proceedings.neurips.cc/paper/2020/file/eae15aabaa768ae4a5993a8a4f4fa6e4-Paper.pdf}
  {Movement pruning: Adaptive sparsity by fine-tuning}.
\newblock In \emph{Advances in Neural Information Processing Systems},
  volume~33, pages 20378--20389.

\bibitem[{Schwartz et~al.(2020)Schwartz, Dodge, Smith, and
  Etzioni}]{schwartz_green_2020}
Roy Schwartz, Jesse Dodge, Noah~A. Smith, and Oren Etzioni. 2020.
\newblock \href {https://doi.org/10.1145/3381831} {Green {AI}}.
\newblock \emph{Communications of the ACM}, 63(12):54--63.

\bibitem[{Shen et~al.(2020)Shen, Dong, Ye, Ma, Yao, Gholami, Mahoney, and
  Keutzer}]{shen2020q}
Sheng Shen, Zhen Dong, Jiayu Ye, Linjian Ma, Zhewei Yao, Amir Gholami,
  Michael~W Mahoney, and Kurt Keutzer. 2020.
\newblock Q-bert: Hessian based ultra low precision quantization of bert.
\newblock In \emph{Proceedings of the AAAI Conference on Artificial
  Intelligence}, volume~34, pages 8815--8821.

\bibitem[{Strubell et~al.(2019)Strubell, Ganesh, and
  McCallum}]{strubell_energy_2019}
Emma Strubell, Ananya Ganesh, and Andrew McCallum. 2019.
\newblock \href {http://arxiv.org/abs/1906.02243} {Energy and {Policy}
  {Considerations} for {Deep} {Learning} in {NLP}}.
\newblock \emph{arXiv:1906.02243 [cs]}.
\newblock ArXiv: 1906.02243.

\bibitem[{Sun et~al.(2019)Sun, Cheng, Gan, and Liu}]{sun2019patient}
Siqi Sun, Yu~Cheng, Zhe Gan, and Jingjing Liu. 2019.
\newblock Patient knowledge distillation for bert model compression.
\newblock In \emph{Proceedings of the 2019 Conference on Empirical Methods in
  Natural Language Processing and the 9th International Joint Conference on
  Natural Language Processing (EMNLP-IJCNLP)}, pages 4323--4332.

\bibitem[{Sun et~al.(2021)Sun, Wang, Feng, Ding, Pang, Shang, Liu, Chen, Zhao,
  Lu, Liu, Wu, Gong, Liang, Shang, Sun, Liu, Ouyang, Yu, Tian, Wu, and
  Wang}]{sun_ernie_2021}
Yu~Sun, Shuohuan Wang, Shikun Feng, Siyu Ding, Chao Pang, Junyuan Shang,
  Jiaxiang Liu, Xuyi Chen, Yanbin Zhao, Yuxiang Lu, Weixin Liu, Zhihua Wu,
  Weibao Gong, Jianzhong Liang, Zhizhou Shang, Peng Sun, Wei Liu, Xuan Ouyang,
  Dianhai Yu, Hao Tian, Hua Wu, and Haifeng Wang. 2021.
\newblock \href {http://arxiv.org/abs/2107.02137} {{ERNIE} 3.0: {Large}-scale
  {Knowledge} {Enhanced} {Pre}-training for {Language} {Understanding} and
  {Generation}}.
\newblock \emph{arXiv:2107.02137 [cs]}.
\newblock ArXiv: 2107.02137.

\bibitem[{Sun et~al.(2020)Sun, Yu, Song, Liu, Yang, and
  Zhou}]{sun-etal-2020-mobilebert}
Zhiqing Sun, Hongkun Yu, Xiaodan Song, Renjie Liu, Yiming Yang, and Denny Zhou.
  2020.
\newblock \href {https://doi.org/10.18653/v1/2020.acl-main.195}
  {{M}obile{BERT}: a compact task-agnostic {BERT} for resource-limited
  devices}.
\newblock In \emph{Proceedings of the 58th Annual Meeting of the Association
  for Computational Linguistics}, pages 2158--2170, Online. Association for
  Computational Linguistics.

\bibitem[{Suresh and Guttag(2021)}]{suresh_framework_2021}
Harini Suresh and John Guttag. 2021.
\newblock \href {https://doi.org/10.1145/3465416.3483305} {A {Framework} for
  {Understanding} {Sources} of {Harm} throughout the {Machine} {Learning}
  {Life} {Cycle}}.
\newblock In \emph{Equity and {Access} in {Algorithms}, {Mechanisms}, and
  {Optimization}}, {EAAMO} '21, pages 1--9, New York, NY, USA. Association for
  Computing Machinery.

\bibitem[{Turc et~al.(2019)Turc, Chang, Lee, and
  Toutanova}]{turc_well-read_2019}
Iulia Turc, Ming-Wei Chang, Kenton Lee, and Kristina Toutanova. 2019.
\newblock \href {http://arxiv.org/abs/1908.08962} {Well-{Read} {Students}
  {Learn} {Better}: {On} the {Importance} of {Pre}-training {Compact}
  {Models}}.
\newblock \emph{arXiv:1908.08962 [cs]}.
\newblock ArXiv: 1908.08962.

\bibitem[{Voita et~al.(2019)Voita, Talbot, Moiseev, Sennrich, and
  Titov}]{voita_analyzing_2019}
Elena Voita, David Talbot, Fedor Moiseev, Rico Sennrich, and Ivan Titov. 2019.
\newblock \href {http://arxiv.org/abs/1905.09418} {Analyzing {Multi}-{Head}
  {Self}-{Attention}: {Specialized} {Heads} {Do} the {Heavy} {Lifting}, the
  {Rest} {Can} {Be} {Pruned}}.
\newblock \emph{arXiv:1905.09418 [cs]}.
\newblock ArXiv: 1905.09418.

\bibitem[{Wang et~al.(2019{\natexlab{a}})Wang, Hula, Xia, Pappagari, McCoy,
  Patel, Kim, Tenney, Huang, Yu, Jin, Chen, Van~Durme, Grave, Pavlick, and
  Bowman}]{wang-etal-2019-tell}
Alex Wang, Jan Hula, Patrick Xia, Raghavendra Pappagari, R.~Thomas McCoy, Roma
  Patel, Najoung Kim, Ian Tenney, Yinghui Huang, Katherin Yu, Shuning Jin,
  Berlin Chen, Benjamin Van~Durme, Edouard Grave, Ellie Pavlick, and Samuel~R.
  Bowman. 2019{\natexlab{a}}.
\newblock \href {https://doi.org/10.18653/v1/P19-1439} {Can you tell me how to
  get past sesame street? sentence-level pretraining beyond language modeling}.
\newblock In \emph{Proceedings of the 57th Annual Meeting of the Association
  for Computational Linguistics}, pages 4465--4476, Florence, Italy.
  Association for Computational Linguistics.

\bibitem[{Wang et~al.(2019{\natexlab{b}})Wang, Singh, Michael, Hill, Levy, and
  Bowman}]{wang2019glue}
Alex Wang, Amanpreet Singh, Julian Michael, Felix Hill, Omer Levy, and
  Samuel~R. Bowman. 2019{\natexlab{b}}.
\newblock {GLUE}: A multi-task benchmark and analysis platform for natural
  language understanding.
\newblock In the Proceedings of ICLR.

\bibitem[{Wang et~al.(2020)Wang, Wohlwend, and Lei}]{wang_structured_2020}
Ziheng Wang, Jeremy Wohlwend, and Tao Lei. 2020.
\newblock \href {https://doi.org/10.18653/v1/2020.emnlp-main.496} {Structured
  {Pruning} of {Large} {Language} {Models}}.
\newblock \emph{Proceedings of the 2020 Conference on Empirical Methods in
  Natural Language Processing (EMNLP)}, pages 6151--6162.
\newblock ArXiv: 1910.04732.

\bibitem[{Yu et~al.(2020)Yu, Edunov, Tian, and Morcos}]{yu_playing_2020}
Haonan Yu, Sergey Edunov, Yuandong Tian, and Ari~S. Morcos. 2020.
\newblock \href {http://arxiv.org/abs/1906.02768} {Playing the lottery with
  rewards and multiple languages: lottery tickets in {RL} and {NLP}}.
\newblock \emph{arXiv:1906.02768 [cs, stat]}.
\newblock ArXiv: 1906.02768.

\bibitem[{Zafrir et~al.(2019)Zafrir, Boudoukh, Izsak, and
  Wasserblat}]{zafrir2019q8bert}
Ofir Zafrir, Guy Boudoukh, Peter Izsak, and Moshe Wasserblat. 2019.
\newblock Q8bert: Quantized 8bit bert.
\newblock \emph{arXiv preprint arXiv:1910.06188}.

\bibitem[{Zhang et~al.(2020)Zhang, Hou, Yin, Shang, Chen, Jiang, and
  Liu}]{zhang2020ternarybert}
Wei Zhang, Lu~Hou, Yichun Yin, Lifeng Shang, Xiao Chen, Xin Jiang, and Qun Liu.
  2020.
\newblock Ternarybert: Distillation-aware ultra-low bit bert.
\newblock In \emph{Proceedings of the 2020 Conference on Empirical Methods in
  Natural Language Processing (EMNLP)}, pages 509--521.

\end{thebibliography}
\bibliographystyle{acl_natbib}

\clearpage
\appendix

\renewcommand\thefigure{A\arabic{figure}}
\section{Experiment Details}
\setcounter{figure}{0} 
\label{sec:experiments}

\subsection{Experiment Breakdown}

\begin{table*}[!h]
    \centering
    \begin{tabular}{lccccc}
    \toprule
    \textbf{Architecture}      & \textbf{\# Layers}    & \textbf{Hidden Dim.}    & \textbf{Params (Millions)}           & \textbf{Size (MB)}      & \textbf{Avg GLUE} \\
    \midrule        
    \textsc{Large}    & 24        & 1024        & 367         & 1341     & 88.47 \\
    \textsc{Base}       & 12         & 768         & 134         & 438     & 87.39 \\
    \textsc{Medium}     & 8         & 512         & 57         & 166      & 85.11 \\
    \textsc{Small}   & 4         & 512         & 45    & 115      & 83.29 \\ 
    \textsc{Mini}     & 4         & 256         & 19         & 45      & 81.41      \\
    \textsc{Tiny}   & 2         & 128         & 8    & 18      & 78.10      \\
    \bottomrule 
    \end{tabular}
    \caption{Information on the different BERT architecture sizes we use in our experiments, with pretrained versions of each size downloaded from \citep{turc_well-read_2019} (in accordance with their license). The ``Avg GLUE'' column is the mean GLUE accuracy across the eight tasks included in our experiments.}
    \label{tbl:archs}
\end{table*}

In total, we tested $576$ experimental conditions, each of which involves fine-tuning a model on a GLUE task. 
We used eight GLUE tasks: \textsc{SST-2, MRPC, STS-B, QQP, MNLI, MNLI-MM, QNLI}, and \textsc{RTE}. 
We excluded \textsc{CoLA} and \textsc{WNLI} from the pruning experiments to reduce the computational burden (more on the compute budget below), and because there are some known issues with \textsc{WNLI} that make it difficult for a fair evaluation \citep{wang-etal-2019-tell}.

We used six architecture sizes (details about the architectures are in Table \ref{tbl:archs}), and eight subsets of compression methods (including the baseline of no compression). 
Note that, when we combine compression methods, there is no concept of order, because all methods function simultaneously and independently: \textsc{QAT} simply adds additional operations after each Linear layer, \textsc{KD} only modifies the loss, and \textsc{MP} gradually masks more weights. 
Therefore, these eight subsets are exhaustive.

There were twice as many pruning experiments as non-pruning experiments, since each pruning experiment tested two different sparsity levels\footnote{Actually, we tested three sparsities (also including $80\%$), but we only show experiments from $40\%$ and $60\%$ in the main text. 
This was because $80\%$ sparse models generally performed poorly, and fell outside the accuracy-model size frontier, so they did not affect our results -- which focused on the best possible tradeoffs for each method. We show these additional results in Appendix \ref{sec:prune}.}. 
So, there were 4 compression subsets without pruning, 4 compression subsets with $40\%$ pruning, and 4 compression subsets with $60\%$ pruning.
Overall, there were $6 \cdot 8 \cdot (3 \cdot 4) = 576$ experimental conditions.

\subsection{Training \& Hyperparameters}

For each experiment, we started by initializing the BERT architecture with pretrained weights from \citet{turc_well-read_2019}. 
We fine-tuned for 3 epochs (on either the base or augmented GLUE training set, depending on whether we were performing a distillation experiment).
As is standard for BERT fine-tuning on GLUE, batch size and LR can have a significant effect on the results \citep{devlin_bert_2018, turc_well-read_2019}. For each experimental condition, we tested three batch sizes ($\{\text{8, 16, 32}\}$) and four learning rates ($\{\text{1e-5, 2e-5, 3e-5, 4e-5}\}$ for \textsc{Large} and \textsc{Base}; $\{\text{3e-5, 5e-5, 0.0001, 0.0003}\}$ for \textsc{Medium}/\textsc{Small}/\textsc{Mini}/\textsc{Tiny}).
For each hyperparameter combination, we performed five repetitions, so there were $3\cdot4\cdot5 = 60$ total training runs per experimental condition. We report max accuracy across these $60$ runs rather than taking an average, as the BERT training on GLUE can be unstable and lead to poor results a high fraction of the time \citep{devlin_bert_2018}. We use the public GLUE development sets rather than the official test sets, since it wouldn't have been feasible to make thousands of submissions to the GLUE testing portal.

Overall, then, we performed $576 \cdot 60 = 34560$ fine-tuning experiments. This was feasible because the smaller architectures could be fine-tuned quickly (from an hour for \textsc{Medium} to a few minutes for \textsc{Tiny}, on the largest GLUE tasks). We performed all experiments on NVIDIA V100 GPUs, and all told, we would estimate approximately 75K GPU hours were necessary for our experiments. As we set a rough budget of 100K GPU hours, this was the reason why we had to make decisions like excluding two GLUE tasks (CoLA, WNLI), not performing data augmentation for non-distillation experiments, and not performing pretraining distillation; any of these decisions would have ballooned our experimental burden. We recognize the privilege of having had access to as much GPU time as we did, and hope that other researchers can benefit from this thorough empirical analysis.

\section{Task-Specific Results}

\begin{figure*}[!ht]
    \centering
    \includegraphics[width=\textwidth]{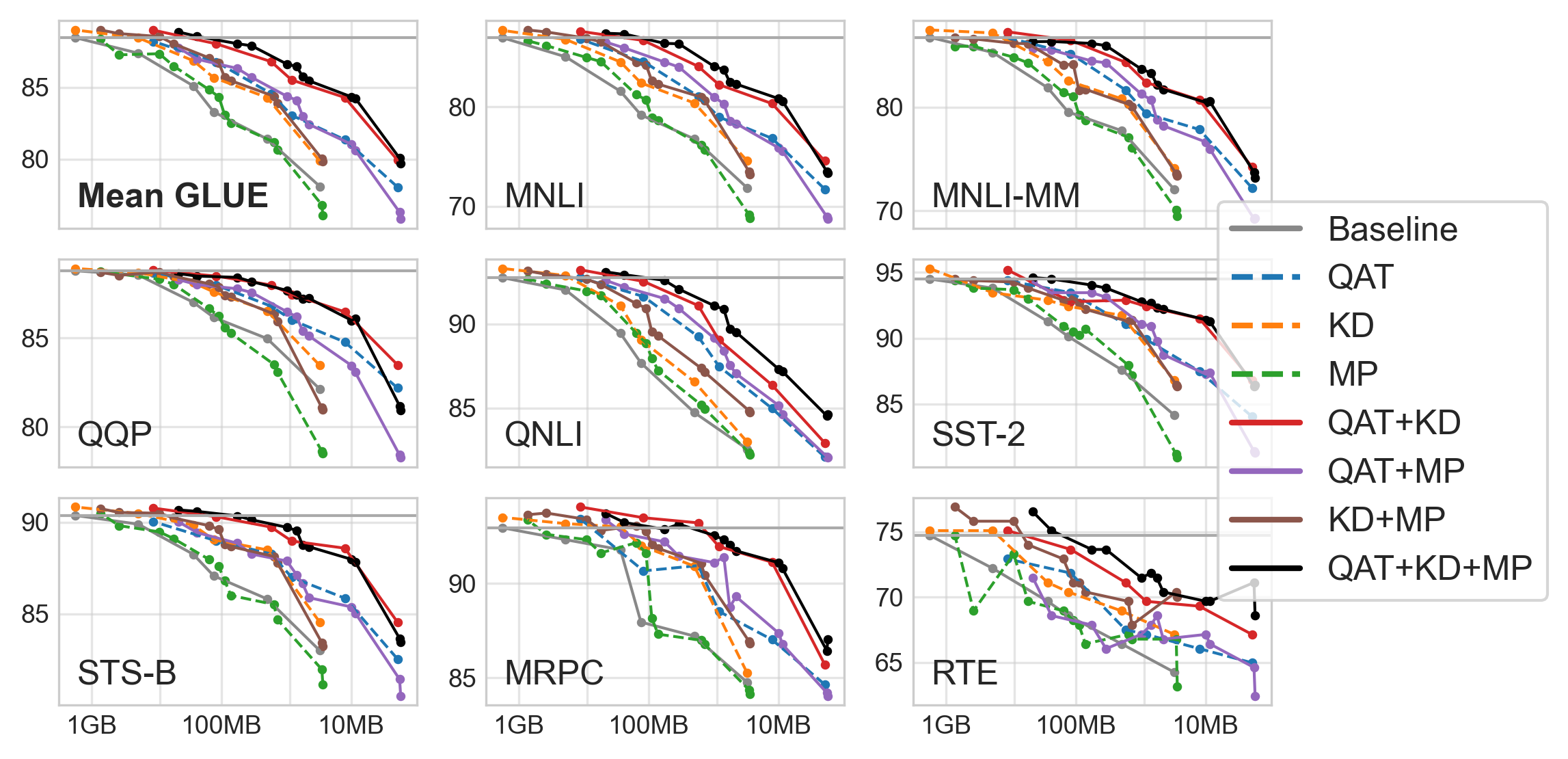}
    \caption{For each task, dev set accuracy vs. decreasing model size (from >1GB down to 10MB), with curves plotted for each compression combination. Tasks are ordered by training dataset size (left-to-right). The grey horizontal line in each plot is the baseline accuracy for BERT-\textsc{Large}.}
    \label{fig:allglue}
\end{figure*}

In Figure \ref{fig:allglue}, we plot similar curves to Figure \ref{fig:meanglue}, split by each task. 
The data from Figure \ref{fig:meanglue} are replicated in the top-left plot. Tasks are ordered by size (reading left-to-right and then top-to-bottom). Most tasks have concordant trends with the curves for average GLUE performance as discussed in the main text, but there are a couple exceptions.

First, for magnitude pruning, we find that some tasks experience \textit{worse} accuracy-size tradeoffs than baseline when pruned to $40\%$ of weights remaining; this was typically not the case when all task accuracies are averaged. Specifically, for some tasks (\textit{e.g.} \textsc{SST-2}, \textsc{STS-B}, \textsc{MRPC}), pruning significantly degrades accuracy below baseline. Smaller BERT architectures are especially harmed, since they are already under-parameterized compared to BERT-\textsc{Large} and \textsc{Base}. This finding agrees with \citet{NEURIPS2020_b6af2c97}, who find that these particular GLUE tasks are least prunable while preserving accuracy (they also use magnitude pruning, but a more compute-intensive version, allowing for slightly higher sparsities than we report). 

Second, we find that for \textsc{MRPC} and \textsc{RTE}, the curves appear noisier, making some trends hard to discern. This is for two reasons: one, the tasks have the smallest training dataset, so they tend to degrade more in response to pruning \citep{NEURIPS2020_b6af2c97}. Second, we empirically found that these tasks varied more in their accuracy from run-to-run than other tasks (perhaps also because of their smaller training datasets). Thus, the true trends for different compression methods may be obscured by lower-confidence accuracy metrics.

\section{Implementation Details}

\subsection{Quantization-Aware-Training}

In this work, we use quantization-aware-training (QAT) rather than naive post-training quantization (PTQ). 
PTQ quantizes weights after training and can significantly increase error due to the loss of precision.
Recently, QAT has been more common, in which the effects of weight quantization are simulated during training with fake quantization operations \citep{jacob_quantization_2017}.
Therefore, at inference time, the model's weights are better tailored to accommodate a reduced precision.

We specifically quantize the embedding and linear modules in our BERT architecture to use INT8 weights, following the symmetric linear quantization scheme from Q8BERT \citep{zafrir2019q8bert}.
The following quantization operation is applied to weights and activations, with scaling factor $S$ and max value $M$, to quantize a value $x$:
$$\mathrm{Quantize}(x \mid S, M) = \mathrm{Clamp}(\lfloor x \cdot S \rceil, -M, M),$$ where $\lfloor \cdot \rceil$ is the integer rounding function, and $\mathrm{Clamp}(\cdot, -M, M)$ maps out-of-range values to $-M$ or $M$. 
$M$ is determined by the number of bits; with 8 bits, for example, we have up to 256 possible quantization levels, so $M = 127$. 
Following \citet{zafrir2019q8bert}, the scaling factor $S$ is set so that the largest possible value for a weight or activation matrix gets quantized to $M$. 
Thus, for a weight matrix $W$, the scale $S^W$ is given by $$S^W = \frac{M}{\max |W|}.$$ 
For activations $x$ from a given layer $L$, the scale factor $S^x$ is computed as an exponential moving average of the max activation value during training, $$S^x = \frac{M}{\mathrm{EMA}(\max_L |x|)}.$$

For quantization-aware-training, we add fake quantization ops to the model's weight matrices and activations during the training forward pass, therefore simulating the effect of quantization on each layer's output. 
However, $\mathrm{Quantize}(\cdot)$ is not differentiable due to the rounding operation, so the backward pass simply ignores the quantization op using the straight-through-estimator: $\partial x^q / \partial x = \vec{\mathbf{1}}$. We model our fake quantization ops off the implementation in Intel's \texttt{nlp-architect}\footnote{\url{https://github.com/IntelLabs/nlp-architect}} repository, authored by \citet{zafrir2019q8bert} and others.

\subsection{Knowledge Distillation}

Knowledge Distillation (KD) aims to transfer the knowledge from a large teacher model into a smaller student model: ideally, our student's predictions will emulate the teacher's, but with reduced compute cost. 
Formally, models trained with KD learn to minimize $\mathcal{L}_{\text{KD}}$, the difference between the teacher's and student's functions $f^T$ and $f^S$, across the training set $\mathcal{X}$: $$\mathcal{L}_{\text{KD}} = \sum_{x \in \mathcal{X}} L\left(f^T(x), f^S(x)\right).$$ 
The functions $f^T(\cdot)$ and $f^S(\cdot)$ include the final output probabilities, and $L(\cdot)$ measures the cross entropy between the student and teacher predicted probabilities \citep{sanh2019distilbert}. 

While some approaches perform distillation during BERT pretraining, we only distill during task fine-tuning, which is also common.
Focusing on fine-tuning was necessary to make our experimental search space tractable, since BERT pretraining can take multiple orders of magnitude more compute than fine-tuning.
Task-specific distillation is also more critical to preserving accuracy than pretraining distillation \citep{jiao2020tinybert}. 
We follow \citet{jiao2020tinybert} in augmenting the GLUE datasets by copying examples and replacing words with synonyms. By running an ablation, they find that augmentation is useful for all tasks, and especially ones with less data (\textit{i.e.,} \textsc{CoLA} and \textsc{MRPC} benefit much more than \textsc{MNLI}). They use different augmentation factors for each dataset, either scaling up the size by 10, 20, or 30 times. We use the same values in our work, copied here: \{\textsc{MNLI}: 10, \textsc{QQP}: 10, \textsc{QNLI}: 20, \textsc{SST-2}: 20, \textsc{STS-B}: 30, \textsc{MRPC}: 30, \textsc{RTE}: 30\}.

We directly used their script, \texttt{data\_augmentation.py}, from the \texttt{TinyBERT}\footnote{\url{https://github.com/yinmingjun/TinyBERT}; no license visible on Github.} Github repository. 
For each GLUE training dataset, this script generates an expanded file in the same format, except with multiple words replaced with synonyms (\textit{i.e.}, L2 nearest neighbors from GLoVe embeddings \citep{pennington-etal-2014-glove}).
We then take our teacher model (fine-tuned BERT-\textsc{Large} on the same GLUE task) and generate predicted probabilities for every example in the augmented dataset, which includes the original sentences and their synonym-replacements.
Note that, when using multiple synonyms, some of the sentences change meaning, leading to a substantially different prediction than the original sentence. 
This is another reason (in addition to little observable change in performance and our compute budget) that we did not use augmentation for the experiments which did not use distillation.

\subsection{Pruning}
\label{sec:prune}

We use the magnitude pruning setup from the \texttt{nn\_pruning} Github repository\footnote{\url{https://github.com/huggingface/nn_pruning}; license allows commercial and private use.} \citep{sanh_movement_2020}. Importantly, because the \texttt{nn\_pruning} implementation of a ``pruned'' model stores the weight values along with a dictionary of weights to be masked, the real model sizes on disk are not smaller. \textbf{So, the pruned model sizes that we report are theoretical rather than actual.} That said, it would be easy to attain a true size reduction if, for example, the weights were changed to zeroes and the model file was gzipped.

As in the movement pruning approach, we prune gradually throughout training. We prune to three possible sparsity levels: $40\%$, $60\%$, or $80\%$ sparsity. Specifically, weights are masked on a linear schedule after 5000 warmup steps, meaning that a constant number of weights are masked at each step in order to reach the target sparsity by the end of training.

For most architectures, the $80\%$ pruning results were very poor: they caused significant accuracy degradation, so they did not yield accuracy-model size tradeoffs that were on the optimal frontier. 
These results did not affect our conclusions, so we removed the $80\%$ sparse points from Figure \ref{fig:meanglue}. We display these results here in the Appendix (Figure \ref{fig:allprune}), by showing the same plots as Figure \ref{fig:allglue}, but with the $80\%$ sparse points added to the curves. There is often a steep accuracy dropoff from $60\%$ to $80\%$ sparsity, especially for the smaller tasks (\textsc{STS-B}, \textsc{MRPC}, \textsc{RTE}).

\begin{figure*}[!h]
    \centering
    \includegraphics[width=\textwidth]{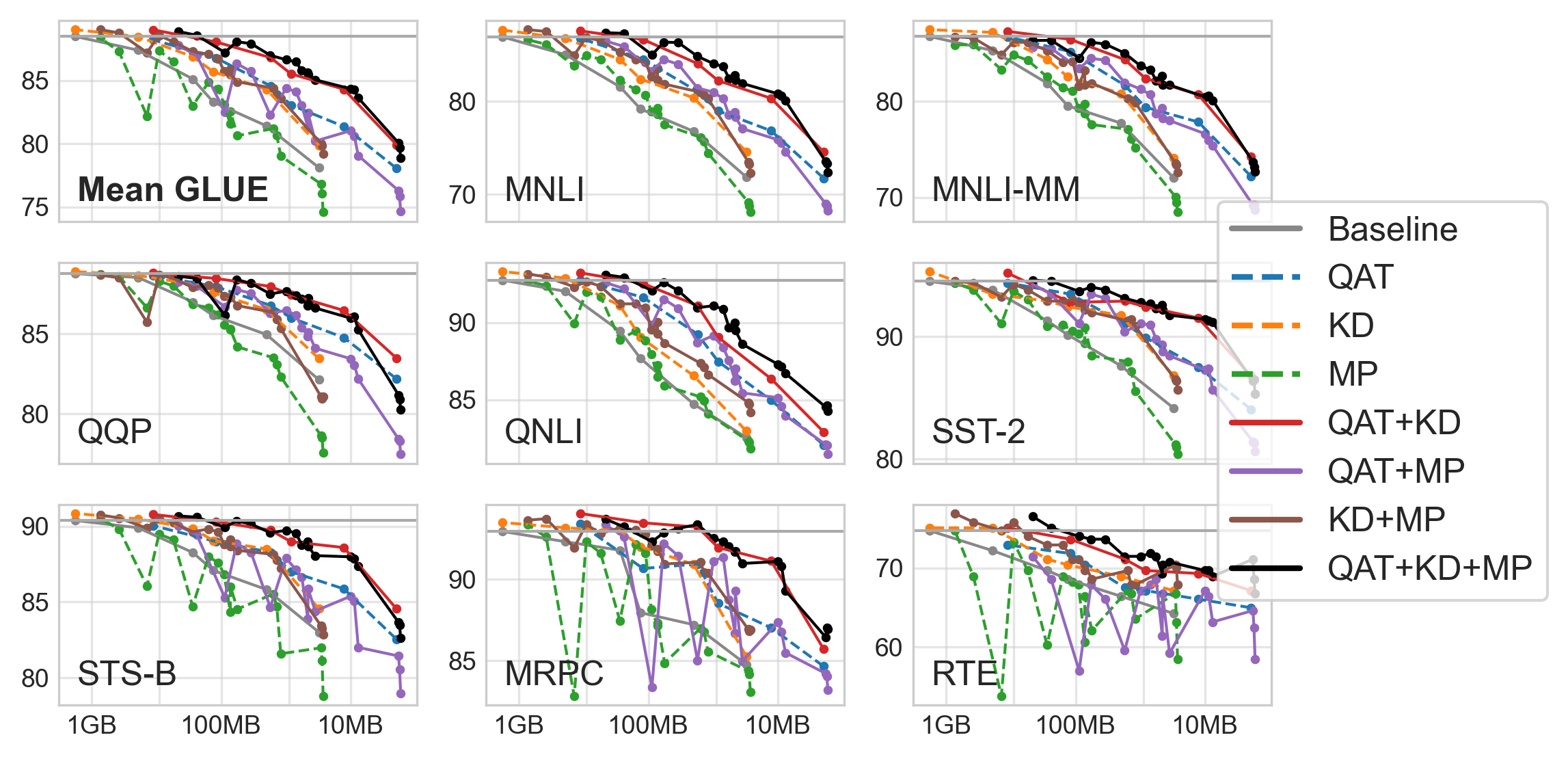}
    \caption{Accuracy vs. decreasing model size; same as Figure \ref{fig:allglue}, but with the $80\%$ pruning experiments also included (\textit{i.e.}, $20\%$ of weights remaining). There is a large dropoff when $80\%$ of weights are pruned compared to $60\%$, especially for smaller tasks.}
    \label{fig:allprune}
\end{figure*}

We acknowledge that there are some marginal improvements on magnitude pruning or other forms of weight pruning that may yield better results for some architectures or tasks.
For example, \citet{NEURIPS2020_b6af2c97} use iterative magnitude pruning with multiple rounds of full training to achieve higher than $40\%$ sparsities without accuracy loss.
However, the goal of our work is not necessarily to achieve the largest model size reductions possible, but rather to understand how methods interact; therefore, we think our conclusions on magnitude pruning would hold even with slight modifications to the method.





\end{document}